\documentclass[letterpaper, 10 pt, conference]{ieeeconf}  
\IEEEoverridecommandlockouts                              
\usepackage{times}
\usepackage{multicol}
\usepackage[bookmarks=true, 
            colorlinks = true,
            linkcolor = blue,
            urlcolor  = blue,
            citecolor = green,
            anchorcolor = green]{hyperref}
            
\usepackage{graphicx}
\usepackage{float} 
\usepackage[font={small,it}]{caption}
\usepackage{subcaption} 
\usepackage[ruled,vlined, noend]{algorithm2e}
\usepackage{booktabs}
\usepackage{multirow}
\usepackage[table, dvipsnames]{xcolor}
\usepackage{amsmath}
\usepackage{dsfont} 
\usepackage{authblk}

\title{\LARGE \bf
Hybrid ICP
}

\author{Kamil Dreczkowski$^{1}$ and Edward Johns$^{1}$
\thanks{$^{1}$The Robot Learning Lab at Imperial College London 
{\tt\small kamil.dreczkowski15@imperial.ac.uk}}%
}
\begin{document}
\maketitle%
\thispagestyle{empty}
\pagestyle{empty}
%
%
\begin{abstract}
%
%
ICP algorithms typically involve a fixed choice of data association method and a fixed choice of error metric. In this paper, we propose Hybrid ICP, a novel and flexible ICP variant which dynamically optimises both the data association method and error metric based on the live image of an object and the current ICP estimate. We show that when used for object pose estimation, Hybrid ICP is more accurate and more robust to noise than other commonly used ICP variants. We also consider the setting where ICP is applied sequentially with a moving camera, and we study the trade-off between the accuracy of each ICP estimate and the number of ICP estimates available within a fixed amount of time.
%
%
\end{abstract}
%
%
\IEEEpeerreviewmaketitle
%
%
\section{Introduction}
%
%
Precise object localisation is essential for robots to interact with their surroundings effectively. The 2020 BOP Challenge \cite{BOPChallenge} revealed the importance of the Iterative Closest Point (ICP) algorithm for accurate pose estimation by showing that 7 out of 10 of the top-performing methods relied on ICP for pose estimation refinement. For example, ICP increased the performance of CosyPose \cite{CosyPose} by $6.1\%$, taking it from 3\textsuperscript{rd} to 1\textsuperscript{st} place in the challenge, and that of Pix2Pose \cite{Pix2Pose} by $24.9\%$, taking it from 22\textsuperscript{nd} to 4\textsuperscript{th} place in the challenge. 
\par
Of all the design choices in an ICP implementation, there are two which affect the accuracy and convergence rate the most \cite{Efficient-variants-of-the-ICP-algorithm}. The first is the data association method used to find correspondences between two point clouds. The second is the error metric minimised when comparing distances between these correspondences. Existing algorithms that use a fixed data association method and fixed error metric commonly fail to estimate object poses robustly, as they are not flexible enough to cope with a range of geometries and ICP initialisations. In this paper, we propose Hybrid ICP, a flexible algorithm that chooses the most appropriate data association method and error metric based on the live image of an object and the current ICP estimate.
\par
\begin{figure}
    \centering
    \includegraphics[width=\linewidth, height=6cm]{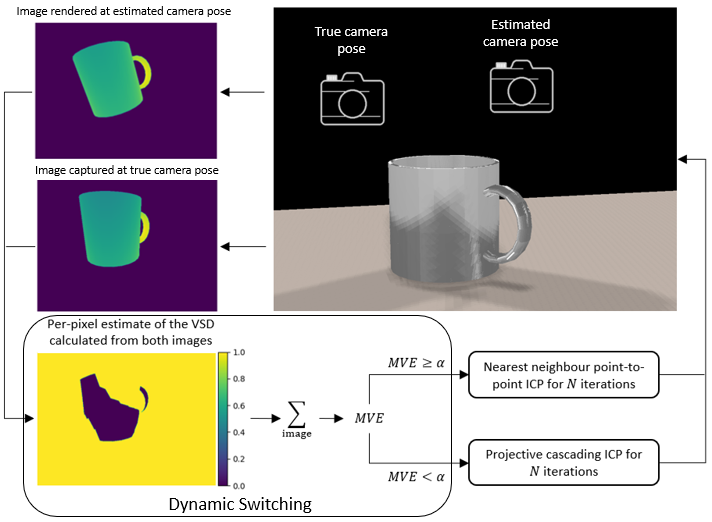}
    \caption{Hybrid ICP is an iterative algorithm that combines Dynamic Switching, Cascading ICP, and point-to-point ICP. A single iteration of Hybrid ICP is shown in this figure. First, the mean VSD estimate (MVE) is calculated using the live image of an object and the current ICP estimate. Then, the MVE is used by Dynamic Switching to optimise the choice of the data association method and error metric. Finally, ICP is used for $N$ iterations to update the current estimate before moving to the next Hybrid ICP iteration.}
    \label{fig:hybrid-icp}
\end{figure}

The contributions of this paper are summarised as follows. For \textbf{our first contribution}, we benchmark some of the most commonly used ICP variants on object pose estimation. This reveals that, and explains why, existing ICP variants fail to estimate the pose of individual objects robustly. For \textbf{our second contribution}, we propose \textbf{Dynamic Switching}, a method that uses an estimate of the  Visible Surface Discrepancy (VSD) \cite{VSD} calculated at the current ICP estimate, which we call the mean VSD estimate (MVE), to optimise the choice of the data association method. For \textbf{our third contribution}, we propose \textbf{Cascading ICP}, which sequentially minimises the point-to-point and then the point-to-plane error metric, to overcome the individual limitations of these error metrics. For \textbf{our fourth contribution}, we introduce the full \textbf{Hybrid ICP}, which combines Dynamic Switching, Cascading ICP, and point-to-point ICP. At each iteration of Hybrid ICP, Dynamic Switching chooses a data association method, which is then combined with either point-to-point or Cascading ICP (see figure \ref{fig:hybrid-icp}). For \textbf{our fifth contribution}, we study how best to fuse multiple individual pose estimates along a trajectory by trading off the accuracy and computational time of each estimate. This is important for applications where sequential observations are available for object pose estimation \cite{coarse-to-fine-imitation-learning, valassakis2021coarse-to-fine}. Supplementary material and videos can be found at \\ \url{https://www.robot-learning.uk/hybrid-icp}.
%
%
%
\section{Related Work}
%
%
The ICP algorithm consists of 5 independent components: 1) \textbf{point selection}, 2) \textbf{data association}, 3) \textbf{correspondence weighting}, 4) \textbf{outlier rejection}, and 5) the \textbf{error metric}. In the past three decades, researchers have proposed many methods for each of these components. For example, \cite{Efficient-variants-of-the-ICP-algorithm, Zippered} propose methods for point selection, \cite{Zippered, Check-normal-compatibility} for outlier rejection, and \cite{Efficient-variants-of-the-ICP-algorithm} outlines methods for weighing correspondences. The main data association methods are nearest neighbour (NN) \cite{point-to-point-icp}, normal shooting \cite{point-to-plane-icp}, and projective \cite{projective-data-association} data association. The main ICP error metrics are the point-to-point \cite{point-to-point-icp} and the point-to-plane \cite{point-to-plane-icp} error metrics. Rusinkiewicz et al. \cite{Efficient-variants-of-the-ICP-algorithm} benchmark many existing ICP variants on three synthetic scenes. As they illustrate, not all components of ICP affect its performance. They found that only the data association method and the choice of the error metric have a large impact, while the remaining components have a data-dependent and/or a negligible impact on performance. Therefore, in contrast to related work, in this paper, we do not propose a novel method for one of the components of ICP, but instead, we propose to use the visible geometry of an object and the current ICP estimate to decide which method to use. The majority of papers that introduce novel ICP variants such as \cite{Generalized-ICP} and \cite{metric-based-icp}, benchmark these variants on either unrealistic or indoor and outdoor scenes. The geometry of such scenes is usually rich in features and is not representative of the geometry of everyday objects. Therefore, even if a variant is shown to perform well on such scenes, it can fail to estimate individual object poses due to the lack of geometric constraints. In this paper, however, we specifically study ICP for the task of individual object pose estimation.
%
%
\section{Background}
%
%
\subsection{Iterative Closest Point (ICP)}\label{sec:ICP}
%
%
The ICP algorithm \cite{point-to-point-icp, point-to-plane-icp} is commonly used for geometric alignment of point clouds when an initial estimate of their transformation is known. When applied to pose estimation, the first point cloud is an object model $O=\{\boldsymbol{o}_i, \boldsymbol{m}_i\}$, and the second point cloud $C=\{\boldsymbol{c}_j, \boldsymbol{n}_j\}$ is computed from a segmented depth image using the camera intrinsic matrix, where each point $\boldsymbol{o}_i$ and $\boldsymbol{c}_j$ is associated with a normal $\boldsymbol{m}_i$ and $\boldsymbol{n}_j$. This paper assumes that segmentation masks are provided by a segmentation method such as the U-Net \cite{U-net}.
\par
\begin{algorithm}
\caption{Generic ICP Algorithm}
\label{alg:standard-icp}
\SetAlgoLined
\DontPrintSemicolon
\KwIn{Point clouds $C=\{\boldsymbol{c}_j, \boldsymbol{n}_j\}$ and $O=\{\boldsymbol{o}_i, \boldsymbol{m}_i\}$\;
      An initial transformation $\boldsymbol{T}_{CO}^1=[\boldsymbol{R}_{CO}^1|\boldsymbol{t}_{CO}^1]$\;
      Maximum number of iterations $max\_iter$}
\KwOut{A refined transformation $\boldsymbol{T}_{CO}$\;}
 \For{$k = 1,..., max\_iter$}{
  \nl $\boldsymbol{c}_i, \boldsymbol{n}_i  \leftarrow \color{Plum}\textbf{DataAssociation}(\boldsymbol{T}_{CO}^{k},  \boldsymbol{o}_i, C)$ for all $i$\;
  \nl $(\boldsymbol{o}_i, \boldsymbol{m}_i) \leftarrow (\boldsymbol{T}_{CO}^k\boldsymbol{o}_i, \boldsymbol{R}_{CO}^k\boldsymbol{m}_i)$ for all $i$\;
  \nl $\boldsymbol{T}_{k} \leftarrow \underset{\boldsymbol{T}}{\arg \min} \underset{i}{\sum}\color{Maroon}\textbf{ ErrorMetric}(\boldsymbol{c}_i, \boldsymbol{n}_i,\boldsymbol{o}_i, \boldsymbol{m}_i, \boldsymbol{T})$\;
  \nl $\boldsymbol{T}_{CO}^{k + 1} \leftarrow \boldsymbol{T}_{k}  \boldsymbol{T}_{CO}^{k}$
 }
\end{algorithm}
Pseudocode of the generic ICP algorithm is shown in Algorithm \ref{alg:standard-icp}. This algorithm only highlights 2 of the 5 components of ICP as the remaining components have been shown to have little effect on its performance \cite{Efficient-variants-of-the-ICP-algorithm}. The algorithm is initialised with two point clouds $C$ and $O$, an initial estimate of the transformation between them, $\boldsymbol{T}_{CO}^1$, and with the maximum number of iterations $max\_iter$. In this work, we assume that ICP initialisation is provided by a global pose estimator such as CosyPose \cite{CosyPose}. During each iteration of the algorithm, a data association method, $\color{Plum}\textbf{DataAssociation}$, assigns points $\boldsymbol{o}_i\in O$ to points $\boldsymbol{c}_j\in C$. All points in $O$ are then transformed using the current ICP estimate. All correspondences $(\boldsymbol{c}_i, \boldsymbol{n}_i, \boldsymbol{o}_i, \boldsymbol{m}_i)$ are then used to minimise an error metric, $\color{Maroon}\textbf{ErrorMetric}$, to solve for an incremental transformation that is used to update the current ICP estimate.
%
%
\subsubsection{\textcolor{Plum}{\textbf{Data Association}}}\label{sec:background:icp:data-association}
%
%
The two most popular data association methods are NN \cite{point-to-point-icp} and projective \cite{ projective-data-association} data association. NN data association uses $\boldsymbol{T}_{CO}^k$ to transform each point $\boldsymbol{o}_i \in O$ into the frame of reference of $C$, and associates each point with its closest neighbour in $C$.
\par
Algorithms that use projective data association store point clouds $C$ and $O$ in vertex maps $V_C, V_O\in\mathds{R}^{H \times W \times 3}$ \cite{KinectFusion}. At each pixel location $(v, u)$, a vertex map stores the $(x, y, z)$ coordinates of the 3D point that was projected to that same pixel in the depth image used to compute the vertex map. $V_O$ is obtained by rendering $O$ at $\boldsymbol{T}_{CO}^1$. Projective data association first uses $\boldsymbol{T}_{CO}^k$ to transform each point $\boldsymbol{o}_i\in V_O$ into the frame of reference of $C$. It then projects that point onto the image plane of $V_C$ using the camera intrinsic matrix. The point $\boldsymbol{c}_j$ stored at the pixel to which $\boldsymbol{o}_i$ is projected is then associated with $\boldsymbol{o}_i$. 
\par 
ICP assumes that each point $\boldsymbol{o}_i \in O$ has a corresponding point $\boldsymbol{c}_i \in C$. As this assumption is often violated, researchers have proposed to only associate points that are no further than $\tau_{max}$ apart \cite{Zippered}, and that have an angle between normals no larger than $\theta_{max}$ \cite{Check-normal-compatibility}. 
%
%
\subsubsection{\textcolor{Maroon}{\textbf{Error metric}}}\label{sec:background:icp:error-metric}
%
%
The two most common error functions are the point-to-point \cite{point-to-point-icp} and point-to-plane \cite{point-to-plane-icp} functions. The point-to-point error function is defined as:
\begin{equation*}
    E_{p2p}(\boldsymbol{c}, \boldsymbol{n}, \boldsymbol{o}, \boldsymbol{m}, \boldsymbol{T}_{CO})= ||\boldsymbol{T}_{CO} \boldsymbol{o} - \boldsymbol{c} ||^2
\end{equation*}
and computes squared distances between correspondences in $O$ and $C$. The point-to-plane error function is defined as:
\begin{equation*}
    E_{p2l}(\boldsymbol{c}, \boldsymbol{n}, \boldsymbol{o}, \boldsymbol{m}, \boldsymbol{T}_{CO})= || \boldsymbol{n}^{T}  ( \boldsymbol{T}_{CO}\boldsymbol{o} - \boldsymbol{c}) ||^2
\end{equation*}
and computes squared distance between points in $O$ and linear surface approximations in $C$.
%
%
\subsection{Visible Surface Discrepancy (VSD)}\label{sec:VSD}
%
%
Evaluating pose estimators that make predictions from single images is challenging, as often more than one object pose is consistent with an input image, and all such poses are equally valid. The VSD \cite{BOPChallenge, VSD} is a pose error function that is invariant to this ambiguity and is defined as:
%
%
\begin{multline}\label{eq:VSD}
    e_{VSD}(D_{est}, D_{gt}, M_{est}, M_{gt}, \tau) =\\ \underset{p \in M_{est} \cup M_{gt}}{\text{avg}} c(p, D_{est}, D_{gt}, M_{est}, M_{gt}, \tau)
\end{multline}
where the cost function is defined as: 
\begin{multline*}
    c(p, D_{est}, D_{gt}, M_{est}, M_{gt}, \tau)\\ = 
    \begin{cases}
    0 \text{ if } p \in M_{est} \cap M_{gt} \wedge |D_{est}(p) - D_{gt}(p)| < \tau \\
    1 \text{ otherwise} 
    \end{cases}
\end{multline*}
where $D_{est}$ and $D_{gt}$ are distance maps obtained by rendering the object at the estimated and ground truth poses, $M_{est}$ and $M_{gt}$ are visibility masks for $D_{est}$ and $D_{gt}$, and $\tau$ is a misalignment tolerance.
%
%
\section{Hybrid ICP}\label{sec:hybrid-icp}
%
%
Hybrid ICP consists of two novel methods: Dynamic Switching, which optimises the data association method, and Cascading ICP, which combines point-to-point and point-to-plane ICP.
%
%
\subsection{Data Association}\label{sec:hybrid-icp:data-association}
%
%
\subsubsection{Limitations of NN and projective data association}
%
%
NN data association relies on distances between points to associate them. For this reason, it is robust to initialisation noise and can nearly always bring two surfaces into better alignment. This is because every point has some corresponding closest point. However, NN algorithms are computationally intensive and are not robust to scenes with surfaces that are very close to each other. To see why this is, consider two neighbouring points on a single surface of a thin object. Given slight misregistration, these points can be matched to points that lie on opposite sides of the object. 
\par
On the other hand, projective data association is performed in constant time and is an order of magnitude faster \cite{Efficient-variants-of-the-ICP-algorithm}. Also, as points are associated based on geometry and the pinhole camera model \cite{CV-modern-approach-chapter-1}, geometric relationships hold between correspondences, resulting in consistent and robust matching. However, for some ICP initialisations, projective data association can fail to find any correspondences altogether. This happens when all points $\boldsymbol{o}\in V_O$ are projected outside of $V_C$ or onto its regions that do not contain the object. 
%
%
\subsubsection{Our solution, Dynamic Switching} \label{sec:method:dynamic-switching}
%
%
We now introduce Dynamic Switching, a novel method that optimises the data association method based on an estimate of the VSD error calculated for the current ICP estimate. This error is an informative proxy for the percentage of points for which projective data association would fail to find a correspondence for. If it is low, we know that many points would be associated as there is good alignment between the input image and an image rendered at the current ICP estimate. In this case, it is best to use projective data association as it yields geometrically consistent correspondences and is faster than NN data association \cite{Efficient-variants-of-the-ICP-algorithm}. However, if the VSD error is large, we can expect that projective data association would find few matches. In this case, it is best to use NN data association as it is more robust to initialisation noise and is more likely to find correspondences.
\par
Dynamic Switching is a simple algorithm that can be integrated into the generic ICP algorithm by using it to optimise the data association method before pose estimation or every $N$ ICP iterations. In essence, Dynamic Switching computes an estimate of the VSD error for the current ICP estimate, and based on its magnitude, decides which data association method to use. If the VSD estimate is more than or equal to some threshold $\alpha$, NN data association is used. And if the estimate is smaller than $\alpha$, projective data association is used. In this project, we have found that $\alpha=0.4$ is a suitable threshold for object pose estimation with noiseless models and images that enable generalisation to unseen objects (see section III of our supplementary material).
\par
From Equation \ref{eq:VSD}, we observe that the VSD error only requires an object's ground truth pose to render a distance map and to compute a visibility mask at that pose. We propose to approximate the distance map and visibility mask with a distance map and segmentation mask of the input depth image. The VSD error estimate calculated using these two approximations is identical to the VSD error in the absence of noise. Similarly to the 2020 BOP Challenge \cite{BOPChallenge} evaluation methodology, we calculate the mean of VSD error estimates for misalignment tolerances ranging from $5\%-50\%$ of an object's diameter, increasing in steps of $5\%$, and refer to this as the \textbf{m}ean \textbf{V}SD \textbf{e}stimate (MVE) hereafter.
%
%
\subsection{Error Metrics}\label{sec:hybrid-icp:error-metrics}
%
%
\subsubsection{Limitations of point-to-point and point-to-plane ICP}
%
%
%
In point-to-point ICP, each point is associated with a discrete pointwise approximation of the other surface. In point-to-plane ICP, each point is associated with a continuous linear approximation of the other surface. The implications of this are twofold. (1) More accurate results can be often obtained with point-to-plane ICP \cite{Check-normal-compatibility}.  When ICP is used for pose estimation, points in $O$ are commonly uniformly distributed across an object's surface, while those in $C$ have a non-uniform distribution as they are computed from an image. Consequently, it is often impossible to precisely align all points from $O$ with those in $C$, while it is possible to align them with continuous linear approximations of the object's surface at these points. (2) Point-to-plane ICP is less robust to simple geometries. The point-to-plane error metric does not penalise points for moving along tangent planes of their correspondences, allowing two surfaces to effectively ``slide" past each other. If the region of overlap between two surfaces being aligned has almost uniform curvature, the two surfaces can diverge starting from a slight misregistration \cite{Check-normal-compatibility}.
%
%
\subsubsection{Our solution, Cascading ICP}\label{sec:method:hybrid-icp:error-metrics:cascading-ICP}
We now introduce Cascading ICP, a two-stage algorithm that combines point-to-point and point-to-plane ICP. During the first stage, Cascading ICP registers two input point clouds with point-to-point ICP. In the second stage, Cascading ICP refines the output of the first stage with point-to-plane ICP. During both stages, if during any ICP iteration:
\begin{enumerate}
    \item the number of correspondences is 0
    \item the number of correspondences is smaller than the number of correspondences during the first or the previous iteration by $5\%$
    \item the mean loss increases compared to the previous iteration
\end{enumerate}
Cascading ICP retrieves the last estimate before the algorithm began diverging and moves to the next stage or outputs the result. This way, if point-to-plane ICP diverges, Cascading ICP returns the output of stage 1. And if both stages diverge, Cascading ICP simply returns ICP initialisation. Both stages are also terminated early if the ICP loss converges.
\par
The reason for refining ICP initialisation with point-to-point ICP before registering two point clouds with point-to-plane ICP is that everyday objects commonly have simple and symmetrical geometries. Refining ICP initialisation decreases the probability of point-to-plane ICP diverging when the two point clouds have few geometric features while enabling Cascading ICP to still benefit from the potentially more accurate result of point-to-plane ICP. 
\par
In preliminary pose estimation experiments, we have found that using Cascading ICP only offers a boost in performance when combined with projective data association. When NN data association is used, Cascading ICP performs consistently worse than point-to-point ICP. This is because everyday objects typically have surfaces that are very close to each other, and NN data association yields many incorrect correspondences for such surfaces. Incorrect correspondences provide misleading constraints for point-to-plane ICP that prevent it from converging to the global minimum. For this reason, Hybrid ICP conditions its choice of error metric on the data association method chosen by Dynamic Switching.
%
%
\subsection{Hybrid ICP - Combining Dynamic Switching, Cascading ICP, and point-to-point ICP} \label{sec:method:hybrid-icp:combining}
%
%
Hybrid ICP is an iterative algorithm that encapsulates the generic ICP algorithm. A single iteration of Hybrid ICP is shown in figure \ref{fig:hybrid-icp}. During each iteration, Hybrid ICP begins by using Dynamic Switching to optimise the choice of the data association method. If projective data association is chosen, Hybrid ICP aligns the two point clouds with Cascading ICP for $N$ iterations. If NN data association is chosen, it aligns the two point clouds with point-to-point ICP for $N$ iterations. 
\par
The intuition behind Hybrid ICP is as follows. During any iteration, if Dynamic Switching chooses projective data association (i.e. the pre-ICP MVE is low), this implies that the two point clouds are already well-aligned. Cascading ICP is the most suitable algorithm in this regime as it is more robust to various geometries than point-to-point or point-to-plane ICP alone (see section III of our supplementary material). If Dynamic Switching chooses NN data association (i.e. the pre-ICP MVE is high), this implies that the two point clouds are not well-aligned and that many found correspondences will be incorrect. In this case, it does not make sense to register the two point clouds with Cascading ICP because normals from incorrect correspondences will provide misleading constraints for the point-to-plane error metric preventing it from aligning the two point clouds. For this reason, we instead use point-to-point ICP to bring the two surfaces into better alignment by insisting that the distance between all correspondences should be minimised. If NN point-to-point ICP successfully aligns the two point clouds, Dynamic Switching will likely choose projective data association in the subsequent Hybrid ICP iteration. In this work, we propose to fix the number of Hybrid ICP iterations to 2 and to allow ICP to converge within each Hybrid ICP iteration.
%
%
%
%
\section{Sequential Pose Estimation}\label{sec:sequential-ICP}
%
%
Consider a camera moving into an static object while sequentially re-estimating its pose. At each time step $t=1,...,T$, the object is at the pose $\boldsymbol{T}_{C_{t}O}$ in the camera's frame, the camera captures an image of the object, and ICP estimates its pose while the camera moves by $\boldsymbol{T}_{C_{t+1}C_{t}}$, where $\boldsymbol{T}_{C_{t+1}O} = \boldsymbol{T}_{C_{t+1}C_{t}} \boldsymbol{T}_{C_{t}O}$. If the camera is mounted to a robot's end-effector, its motion can be accurately determined from robot kinematics and extrinsic camera calibration.
\par
When transitioning from time $t$ to $t+1$, all previous estimates must be transformed by the camera's motion to account for the movement of its frame of reference, i.e. each previous estimate $\boldsymbol{\Tilde{T}}_{C_{t}O}^l$ must be transformed to the new camera frame according to  $\boldsymbol{\Tilde{T}}_{C_{t+1}O}^l = \boldsymbol{T}_{C_{t+1}C_{t}} \boldsymbol{\Tilde{T}}_{C_{t}O}^l$, where the superscript $l$ is the time step at which the estimate was made. Given this, at every time step $t$, we have a list of estimates of the object's pose in the current camera frame, possibly with associated uncertainties i.e. $    \{\boldsymbol{\Tilde{T}}_{C_{t}O}^1, \boldsymbol{\Sigma}^1\}, ..., \{\boldsymbol{\Tilde{T}}_{C_{t}O}^{t-1}, \boldsymbol{\Sigma}^{t-1}\}$, where $\boldsymbol{\Sigma}^l$ is the uncertainty associated with the estimate $\boldsymbol{\Tilde{T}}_{C_{t}O}^l$. In this work, we propose to use the MVE, $e_{MVE}$, calculated for an ICP estimate as an uncertainty proxy. Although not theoretically grounded, the MVE is relatively cheap to compute and broadly captures the accuracy of an estimate. In the next section, we describe 6 methods which can be used to initialise ICP at every time step of a trajectory, and to calculate a final pose estimate given multiple estimates made at previous time steps. The \textbf{Average}, \textbf{Weighted average}, \textbf{Filtering} and \textbf{Filtering with constant weights} methods all use the algorithm proposed by Barfoot et al. \cite{Associating-uncertainty-SE3} for fusing $\textit{SE}(3)$ estimates (see section I of our supplementary material for details). Out of these four methods, the \textbf{weighted average} and \textbf{filtering} methods require an estimate of the covariance associated with each pose estimate. We propose to use $e_{MVE}$ multiplied by the identity matrix $\boldsymbol{I}\in \mathds{R}^{6\times6}$ as a proxy for this covariance. 
%
%
%
%
%
\subsection{Methods for Sequential ICP Initialisation}
%
%
\noindent\textbf{Last Estimate}: When queried at time step $t$, this method returns $\boldsymbol{\Tilde{T}}_{C_{t}O}^{t-1}$.
\par
\noindent\textbf{Average}: When queried for an estimate, this method takes a deterministic average of all previous estimates and returns the result.
\par
\noindent\textbf{Weighted Average}: When queried for an estimate, this method fuses all previous estimates using their associated uncertainties and returns the result.
\par
\noindent\textbf{Filtering with constant weights}:  After ICP makes an estimate $\boldsymbol{\Tilde{T}}_{C_{t}O}^t$, this method fuses $(\boldsymbol{\Tilde{T}}_{C_{t}O}^t, \boldsymbol{I})$ with the previous estimate $(\boldsymbol{\Tilde{T}}_{C_{t}O}^{t-1}, \boldsymbol{\Sigma}^{t-1})$ to obtain a new estimate $(\boldsymbol{\Bar{T}}_{C_{t}O}, \boldsymbol{\Bar{\Sigma}})$. It then sets $(\boldsymbol{\Tilde{T}}_{C_{t}O}^t, \boldsymbol{\Sigma}^t)\leftarrow (\boldsymbol{\Bar{T}}_{C_{t}O}, \boldsymbol{\Bar{\Sigma}})$. When queried for an estimate at time step $t$, this method returns $\boldsymbol{\Tilde{T}}_{C_{t}O}^{t-1}$.
\par
\noindent\textbf{Filtering}: This is identical to the method described above, with the exception that after ICP makes an estimate, it fuses $(\boldsymbol{\Tilde{T}}_{C_{t}O}^t, e_{MVE}^t\boldsymbol{I})$ with $(\boldsymbol{\Tilde{T}}_{C_{t}O}^{t-1}, \boldsymbol{\Sigma}^{t-1})$ to obtain $(\boldsymbol{\Bar{T}}_{C_{t}O}, \boldsymbol{\Bar{\Sigma}})$.
\par
\noindent\textbf{Most Confident}: This method stores all previous measurements with their associated uncertainties. When queried for an estimate, it returns the estimate with the lowest uncertainty.
%
%
\section{Experiments}
%
%
%
%
\subsection{Experimental Procedure}\label{sec:experiments:experimental-procedure}
%
%
This section only briefly outlines the experimental procedure adopted in our simulation experiments. For further details, see section II of our supplementary material.
%
%
\subsubsection{Objects}
%
%
\begin{figure}
    \centering
    \includegraphics[width=\linewidth, height=3.5cm]{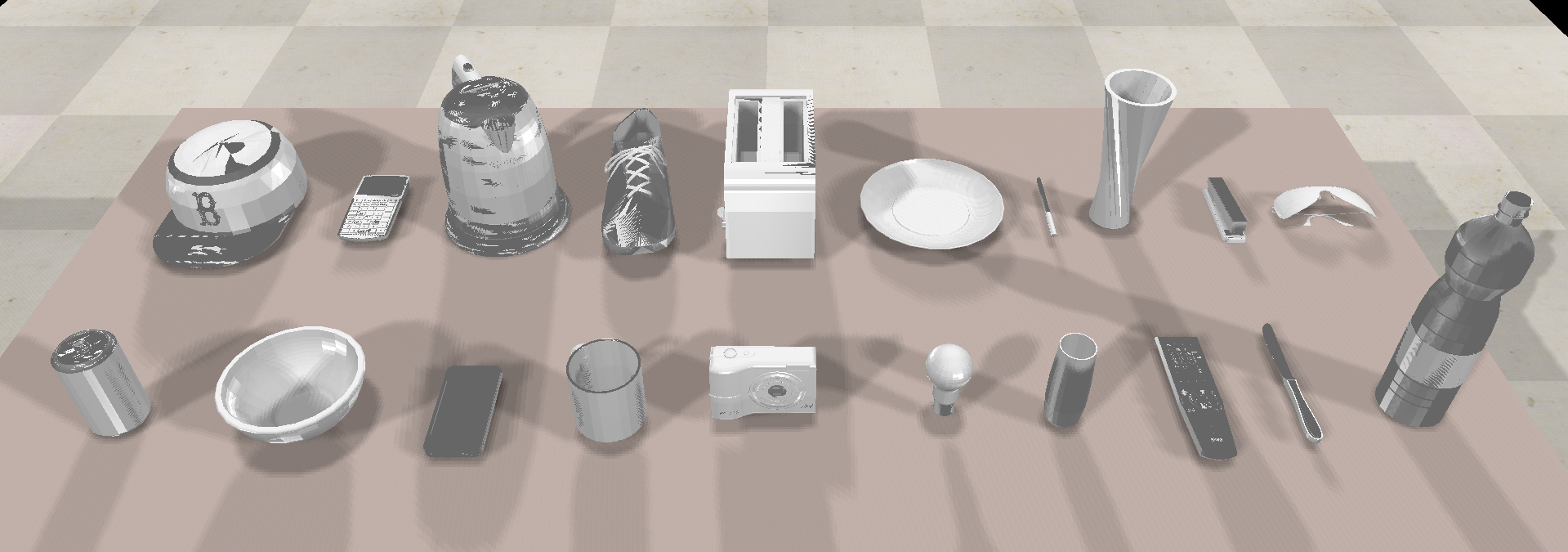}
    \caption{The 20 objects considered in this project. The top row illustrates the 10 objects used to tune the Dynamic Switching threshold. The bottom row illustrates the 10 objects used in our main experiments.}
    \label{fig:experiments:objects-considered}
\end{figure}
%
We consider 20 ShapeNet \cite{shapenet} objects from 20 different object categories. These objects are shown in figure \ref{fig:experiments:objects-considered}. We use the 10 objects shown in the top row of this figure to tune the Dynamic Switching threshold (see section III of our supplementary material), and the remaining 10 to benchmark Hybrid ICP against other algorithms.
%
%
\subsubsection{Evaluation Metric}
%
%
In line with the evaluation methodology of the 2020 BOP Challenge \cite{BOPChallenge}, we compute the VSD pose error for misalignment tolerances ranging from $5\%-50\%$ of an object's diameter, increasing in steps of $5\%$, and report the average of these errors marginalised across objects, object poses, and initialisation poses.
%
%
\subsubsection{Sampling Object Poses}\label{sec:sampling-object-poses}
%
%
In single-image experiments (section \ref{sec:experiments:single-image}),  we sample object poses such that the distance between the object and the camera is uniformly distributed between $[d_{obj}, 0.6]$m, where $d_{obj}$ is the object's diameter.  In trajectory-based experiments (section \ref{sec:experiments:trajectory-based}), we fix the distance between the object and the camera to 1m. In all experiments, we set the camera's orientation so that its optical axis intersects with a point in the proximity of the object's centre. This ensures that an object is never rendered exactly at the centre of an image. 
\subsubsection{Sampling Initialisation Poses}\label{sec:sampling-initialisation-poses}
%
%
Given a magnitude of the translation and rotation perturbation, $\delta_t$ and $\delta_\theta$, we first uniformly sample two vectors $\boldsymbol{v}$ and $\boldsymbol{\eta}$ from the surface of the unit sphere. The translation perturbation is then defined as $\boldsymbol{t}_\delta = \delta_t\boldsymbol{v}$. The rotation perturbation $\boldsymbol{R}_\delta \in \textit{SO}(3)$ is obtained by converting the axis-angle vector $(\delta_\theta, \boldsymbol{\eta})$ to a matrix. We define ICP initialisation as $\boldsymbol{\Tilde{T}}_{CO}^1=[\boldsymbol{R}_{CO} \boldsymbol{R}_\delta| \boldsymbol{t}_{CO} + \boldsymbol{t}_\delta]$, where $\boldsymbol{T}_{CO}=[\boldsymbol{R}_{CO}| \boldsymbol{t}_{CO}]$ is an object's ground truth pose. 
%
%
\subsubsection{Relationship between translation and rotation errors of a sampling-based pose estimator}\label{sec:ratio}
%
%
We first estimated the relationship between the magnitudes of translation and rotation errors of a sampling-based pose estimator to generate realistic initialisations.
From preliminary experiments, we have found that on average, estimated poses had $1.92^{\circ}$ of rotation error per $1$ mm of translation error. We have also found that the mean translation and rotation error was $6.2$mm and $9.5^{\circ}$ respectively.
\subsection{Single Image Pose Estimation Refinement}\label{sec:experiments:single-image}
%
%
\begin{figure*}[t]
    \begin{multicols}{3}
    \includegraphics[width=\linewidth, height=4.2cm]{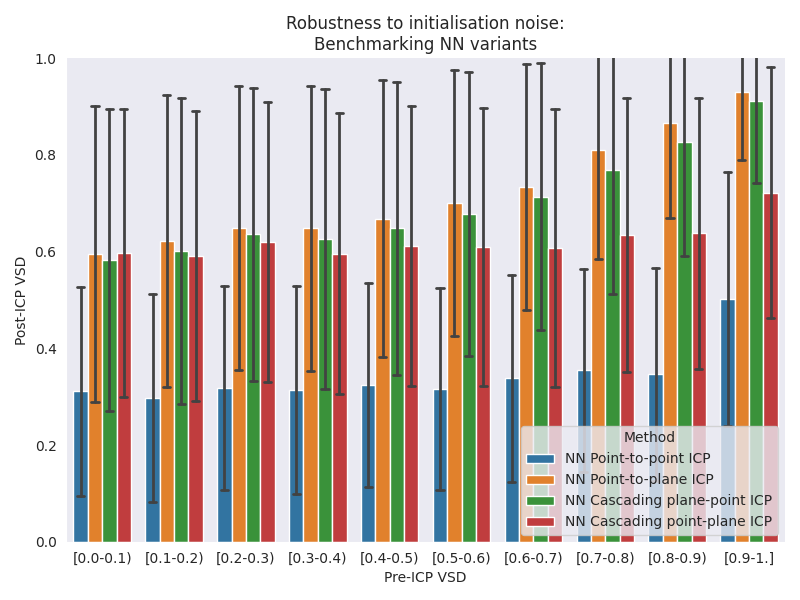}\par 
    \includegraphics[width=\linewidth, height=4.2cm]{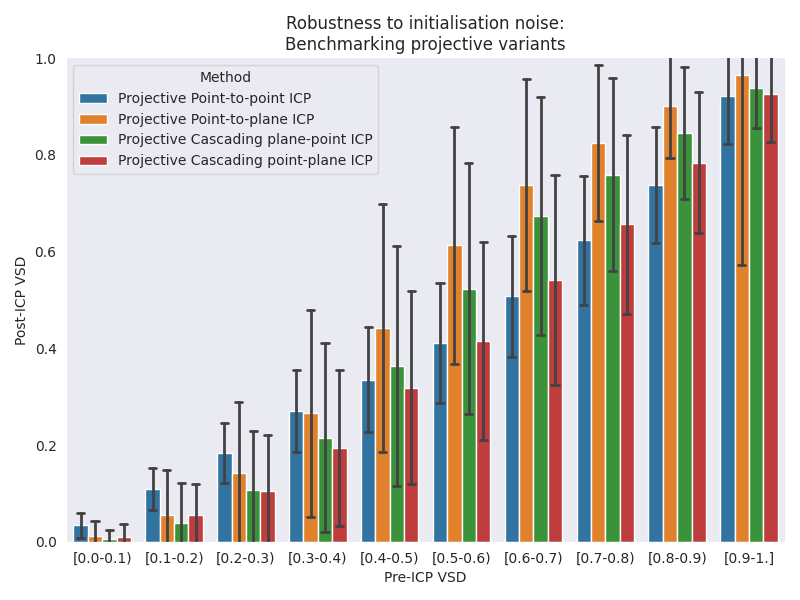}\par 
    \includegraphics[width=\linewidth, height=4.2cm]{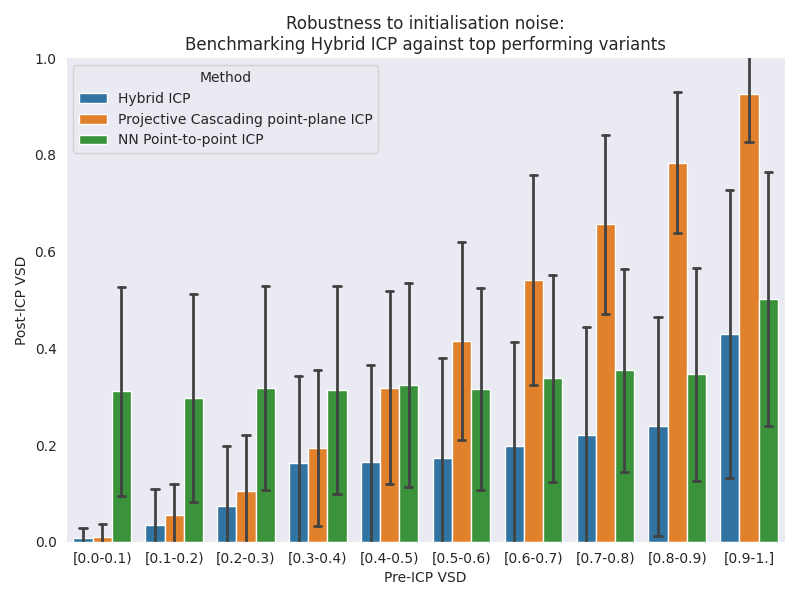}\par
    \end{multicols}
    \begin{multicols}{3}
    \includegraphics[width=\linewidth, height=4.2cm]{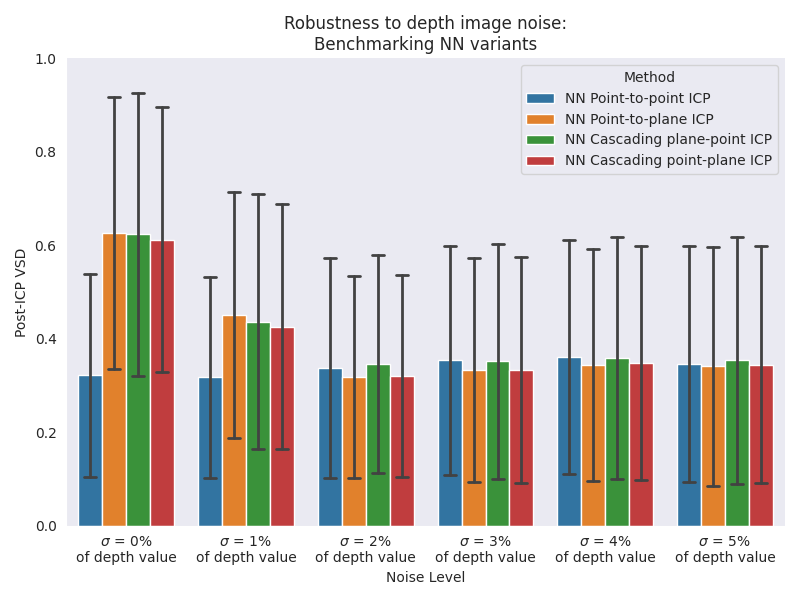}\par 
    \includegraphics[width=\linewidth, height=4.2cm]{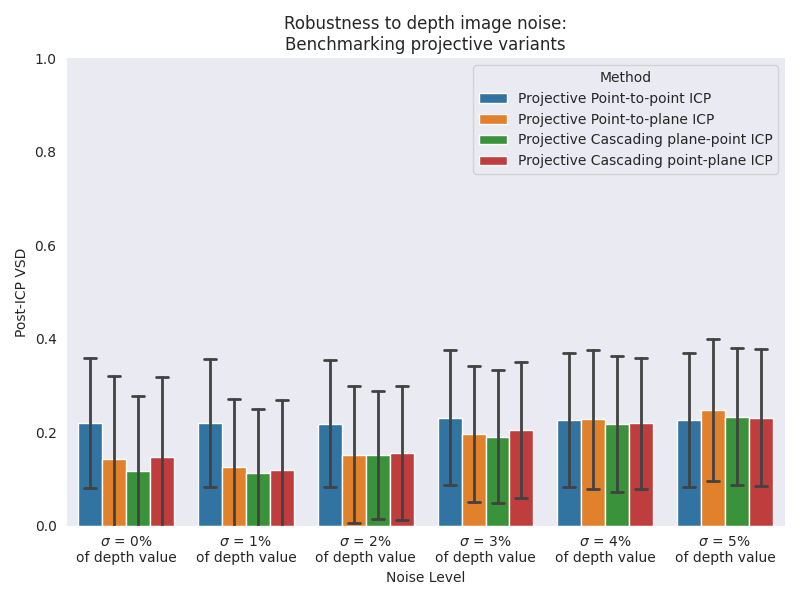}\par 
    \includegraphics[width=\linewidth, height=4.2cm]{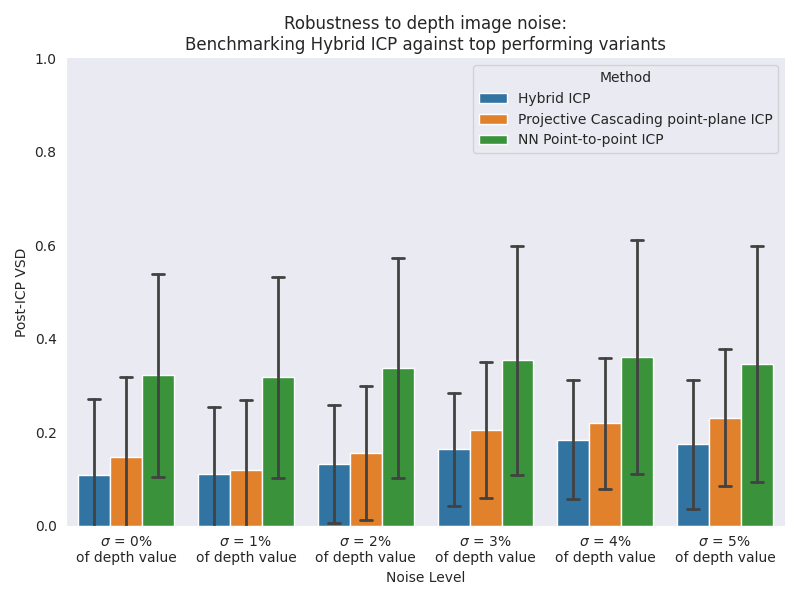}\par
    \end{multicols}
    \begin{multicols}{3}
    \includegraphics[width=\linewidth, height=4.2cm]{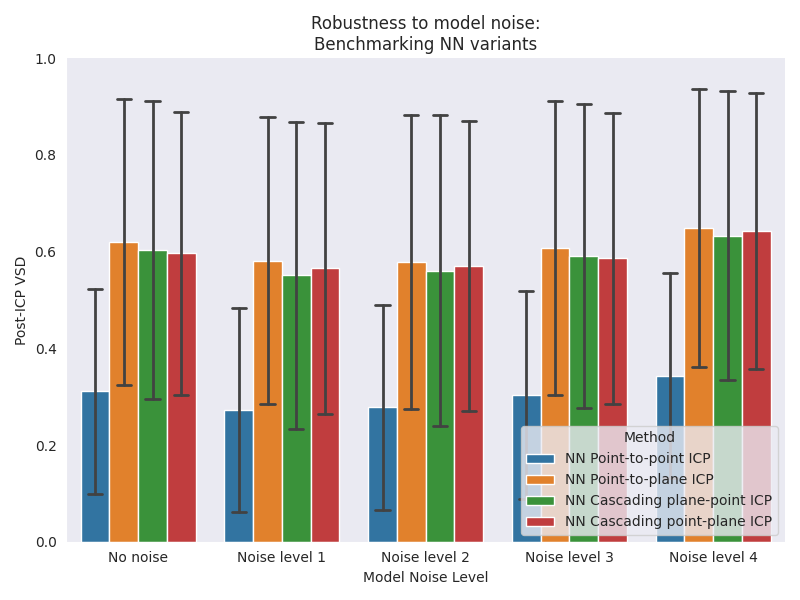}\par 
    \includegraphics[width=\linewidth, height=4.2cm]{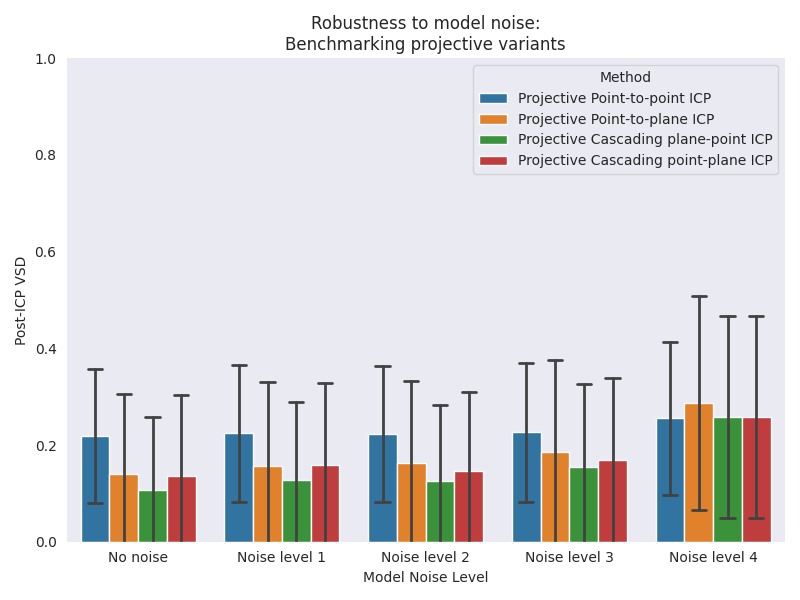}\par 
    \includegraphics[width=\linewidth, height=4.2cm]{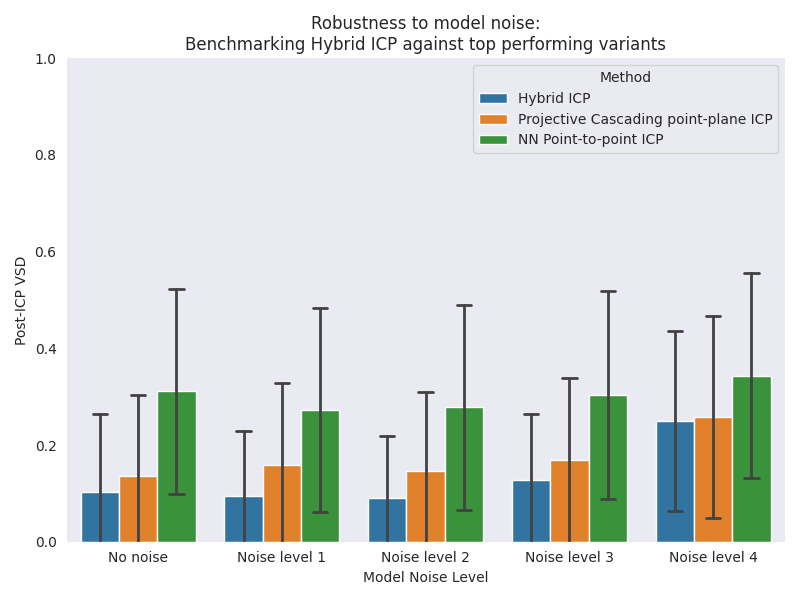}\par
    \end{multicols}
\caption{The first column benchmarks NN ICP variants, the second benchmarks projective ICP variants, and the last benchmarks Hybrid ICP against the most robust NN and projective variant. The top row investigates robustness to initialisation noise, the second row robustness to depth image noise, and the final row robustness to model noise. Pre-ICP VSD is the mean VSD calculated for ICP initialisation. Post-ICP VSD is the mean VSD calculated for the output of ICP. Note that the lower the mean VSD, the more accurate a method is.} \label{fig:experiments:single-image:all-results}
\end{figure*}
We now describe three experiments that were carried out to test the robustness of Hybrid ICP to 1) \textbf{initialisation noise}, 2) \textbf{depth image noise}, and 3) \textbf{model noise}. All main results are illustrated in Figure \ref{fig:experiments:single-image:all-results}, where each row corresponds to each of the three experiments. The first column benchmarks NN variants, the second projective variants, and the final benchmarks Hybrid ICP against the most robust NN and projective variant. As baselines, we implement point-to-point, point-to-plane and Cascading ICP, all with both NN and projective data association. For completeness, we also implement a Cascading ICP variant that first does point-to-plane ICP and then \textit{cascades} to point-to-point ICP. Hereafter, we refer to the original variant (see section \ref{sec:method:hybrid-icp:error-metrics:cascading-ICP}) as \textit{cascading point-plane ICP}, and to this variant as \textit{cascading plane-point ICP}. 
%
%
\subsubsection{Robustness to Initialisation Noise}\label{sec:exp:initialisation}
%
%
We quantify initialisation noise as the mean VSD error calculated for ICP initialisation (pre-ICP VSD). In this experiment, we sample translation perturbations from a uniform distribution, $\delta_t\sim \mathcal{U}[0, 0.15]$ m, and calculate the corresponding rotation perturbations $\delta_\theta$ according to the ratio described in section \ref{sec:ratio}. We use rejection sampling to ensure a uniform distribution of pre-ICP VSD errors across 10 equally spaced bins. In total, we sample 1000 object poses per object. All results for this experiment are shown in the top row of Figure \ref{fig:experiments:single-image:all-results}.
\par
Observing the first two figures, we notice that variants that use a fixed data association method and error metric fail to estimate object poses reliably. This is because these variants are not flexible enough to cope with simple geometries with few features and all possible initialisations. The two figures also reveal that both variants fail in the high pre-ICP VSD regime and that projective variants tend to perform best in the low pre-ICP VSD regime while NN point-to-point ICP in the high pre-ICP VSD regime. This is because when the pre-ICP error is high, projective variants fail to find correspondences, while NN variants do not have this problem. As the rightmost graph illustrates, Hybrid ICP is the most robust variant across all pre-ICP VSD intervals. This figure also shows that Hybrid ICP outperforms projective cascading point-plane ICP even in the low pre-ICP VSD regime. This is because false correspondences found by projective data association can result in an update that worsens the registration. If the region of overlap between the surfaces being aligned does not decrease and the mean loss does not increase, Cascading ICP would continue as normal. In contrast to this, Hybrid ICP would detect this as the MVE would increase, and would switch to NN data association to coarsely re-align the two surfaces.
%
%
\subsubsection{Robustness to Depth Image Noise}
%
%
In this experiment, we assume Gaussian noise in depth images with a standard deviation of $x\%$ of a pixel's depth value, for $x\in[0, 1, 2, 3, 4, 5]$. To evaluate all methods, we sample 100 poses per object and render the objects at these poses. We then perturb each ground truth pose by a translation and rotation perturbation $\delta_t=6.2$ mm and $\delta_\theta =9.5^\circ$ (i.e. by the mean error of the sampling-based pose estimator) to obtain ICP initialisation. Finally, we evaluated all methods using identical depth images and initialisations by adding different noise levels to the original images. Results are shown in the middle row of Figure \ref{fig:experiments:single-image:all-results}. 
\par
Observing the first two figures, we notice that projective ICP variants are less susceptible to noise in depth images. We hypothesise that this is because projective variants yield geometrically consistent correspondences while NN variants are sensitive to noise. Another interesting observation is that the performance of NN and projective point-to-plane, Cascading point-plane, and Cascading plane-point increases after the standard deviation of the noise is increased from $0\%$ to $1\%$. Although counter-intuitive, we hypothesise that the reason for this is that all test objects have simple geometries with few features, while all of these variants rely on point-to-plane ICP. In the absence of noise, these variants are brittle and susceptible to the surfaces ``sliding" past each other during ICP. However, in the presence of noise, noisy normals help to ``lock" the surfaces in place, limiting their ability to ``slide" and diverge. Finally, examining the rightmost figure reveals that Hybrid ICP is the most robust variant to noise in depth images.
%
%
\subsubsection{Robustness to Model Noise}
%
%
\begin{figure}
    \centering
    \includegraphics[width=\linewidth, height=3cm]{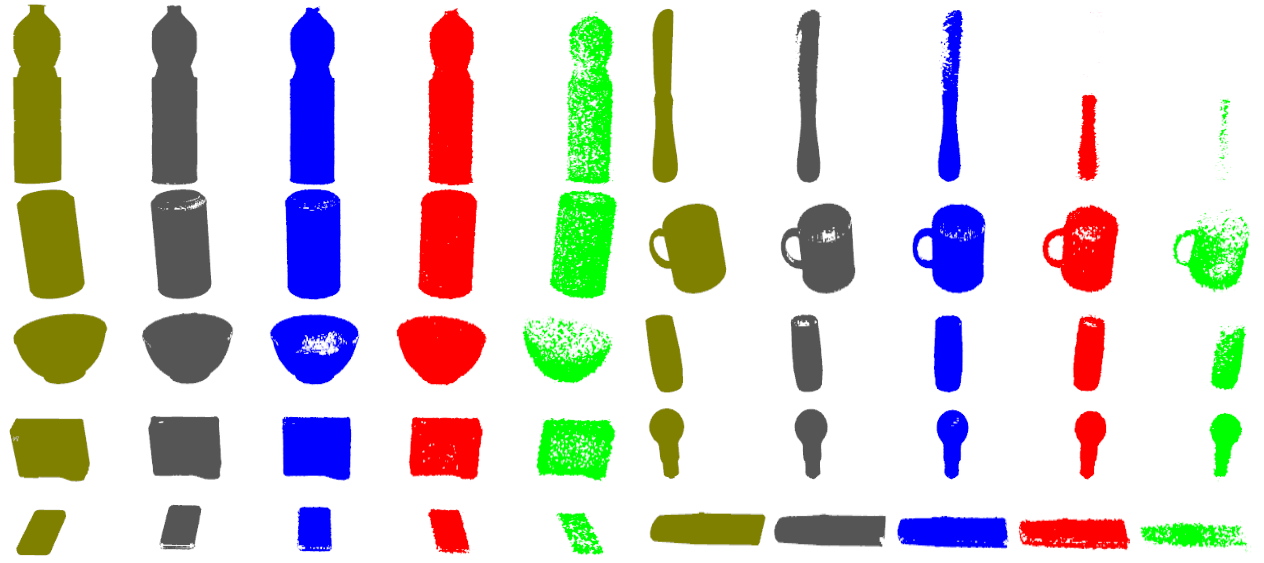}
    \caption{Triangular meshes of object CAD models and reconstructions. Yellow objects correspond to CAD models (``\textit{No Noise}" in Figure \ref{fig:experiments:single-image:all-results}). The grey, blue, red and green objects correspond to reconstructions 1, 2, 3 and 4 respectively (``\textit{Noise level i}" for $i=[1,2,3,4]$ in Figure \ref{fig:experiments:single-image:all-results}).}
    \label{fig:experiments:noisy-models}
\end{figure}
We model noise in object models by creating 4 reconstructions of each of the objects with varying levels of noise with TSDF integration \cite{TSDF-integration}. We create each reconstruction by integrating 100 randomly sampled images. Before integrating images to obtain reconstruction $i\in[1, 2, 3, 4]$, we:
\begin{itemize}
    \item Add $(i-1)/2$ pixels to all camera intrinsic parameters,
    \item Add Gaussian noise to all depth images with a standard deviation equal to $(i-1)\%$ of a pixels depth value,
    \item Perturb all ground truth object poses by a translation and rotation perturbation $\delta_t=(i-1)$ mm and $\delta_\theta=(i-1)^\circ$.
\end{itemize}
%
%
Figure \ref{fig:experiments:noisy-models} illustrates triangular meshes of all objects for all five noise levels. We sample 100 object poses per object to evaluate all variants and render the objects at these poses. We then perturb each ground truth pose by a translation and rotation perturbation $\delta_t=6.2$ mm and $\delta_\theta =9.5^\circ$ (i.e. by the mean error of the sampling-based pose estimator). Finally, we evaluated all methods using identical depth images and initialisations while using object models with different noise levels. All results for this experiment are shown in the bottom row of Figure \ref{fig:experiments:single-image:all-results}.
\par
Once again, the first two figures reveal that projective ICP variants are more robust to model noise than NN variants. Again, we hypothesise that this is because projective variants yield geometrically consistent correspondences. Also, from the rightmost figure, we can see that Hybrid ICP is the most robust variant to model noise. 
%
%
\subsubsection{Speed of Convergence}
%
%
Table \ref{table:Convergence-rate} shows the mean and standard deviation of inference times for all evaluated algorithms, averaged across all three single-image experiments. As these results illustrate, projective variants are significantly faster than Hybrid ICP and NN variants. 
%
\begin{table}[H]
\caption{Inference time of all ICP variants (fastest on top).}

\label{table:Convergence-rate}
\begin{center}
\begin{tabular}{|l|c|}
\hline
\textbf{Method} & \textbf{Execution time (s)} \\
\hline
Projective point-to-plane  & $0.061 \pm 0.023$ \\
\hline
Projective point-to-point  & $0.085 \pm 0.038$ \\
\hline
Projective cascading plane-point & $0.094  \pm 0.035$ \\
\hline
Projective cascading point-plane & $0.101 \pm 0.040$\\
\hline
NN point-to-point & $0.401 \pm 0.516$\\
\hline
Hybrid ICP & $0.542 \pm 0.436$ \\
\hline
NN point-to-plane & $0.786 \pm 3.535$ \\
\hline
NN cascading plane-point & $0.879 \pm 3.713$ \\
\hline
NN cascading point-plane & $1.090 \pm 3.869$ \\
\hline
\end{tabular}
\end{center}
\end{table}
%
%
\subsection{Sequential ICP}\label{sec:experiments:trajectory-based}
%
%
We now consider the setting where ICP is applied sequentially with a camera moving into a static object along a trajectory and refer to this setting as a \textit{Sequential ICP}. Here, we study different ICP variants for their trade-off of accuracy and convergence speed, since faster variants allow for more pose estimations in a given time.
%
%
\subsubsection{Trajectory definition}
%
%
We define a trajectory as a linear path between some initial camera pose and a point in the proximity of an object. The camera pose is sampled in line with the description in section \ref{sec:sampling-object-poses}, and the camera moves in the direction of its optical axis at a velocity of $0.1$ ms\textsuperscript{-1}. The magnitude of the camera movement between any two subsequent time steps is determined by the time it takes a method to obtain ICP initialisation and calculate the uncertainty proxy when required. 
%
%
\subsubsection{Experiment}
%
%
\begin{figure*}
    \begin{multicols}{2}
    \centering
    \includegraphics[width=\linewidth, height=5.5cm]{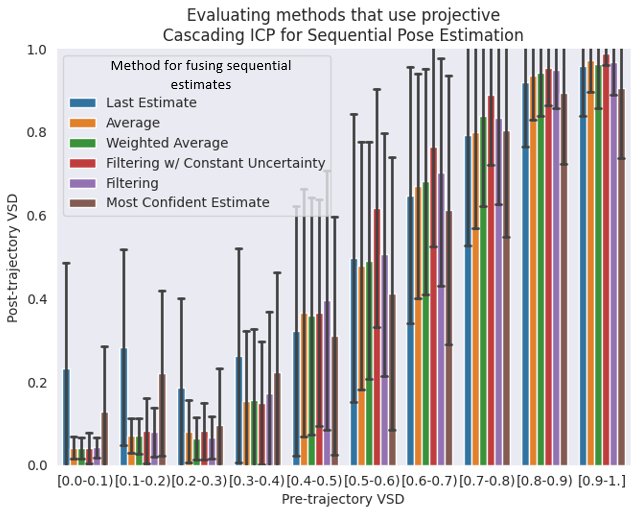}\par 
    \centering
    \includegraphics[width=\linewidth, height=5.5cm]{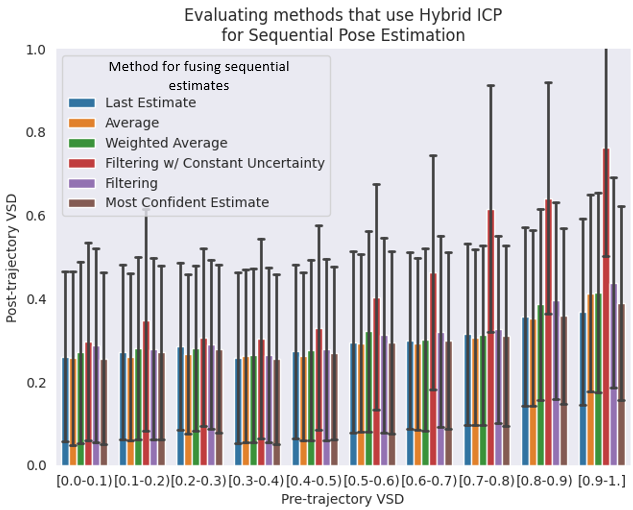}\par
    \end{multicols}
    %
    %
\caption{Experimental results for sequential ICP. The figures on the left and right show the performance of all methods for fusing multiple estimates when using projective Cascading ICP and Hybrid ICP as the underlying ICP algorithm respectively. Note that the lower the post-trajectory VSD the better.} \label{fig:experiments:sequential-icp}
\end{figure*}
To study the trade-off between the accuracy of each ICP estimate and the number of available estimates, we use projective Cascading (point-plane) and Hybrid ICP as the underlying ICP algorithms. Hybrid ICP was shown to consistently perform best in single-image experiments while having a relatively high inference time. On the other hand, projective Cascading ICP was shown to perform well in the low pre-ICP VSD regime and poorly in the high pre-ICP VSD regime while having a low inference time. By using these two variants, we investigate if it is better to have many low-quality estimates or fewer high-quality estimates.
%
%
In line with the single image experiment, to initialise each trajectory, we sample translation perturbations from the uniform distribution $\delta_t\sim \mathcal{U}[0, 0.15]$ m and calculate the corresponding rotation perturbations $\delta_\theta$ according to the ratio described in section \ref{sec:ratio}. We use rejection sampling to ensure a uniform distribution of pre-trajectory VSD errors across 10 equally spaced bins. To simulate a realistic setting, we assume the depth noise model proposed by Ahn et al. \cite{RealSense-noise-model} for the Intel RealSense D435 depth camera and access to a noisy reconstruction.
%
%
%
%
To this end, we reconstruct each of the objects using 100 random images assuming the RealSense D435 noise model and by perturbing ground-truth camera poses by a translation and rotation perturbation $\delta_t=5$ mm and $\delta_\theta=1.1^\circ$. These magnitudes were obtained by calculating the mean error for multiple extrinsic calibrations of a RealSense D435 camera mounted to a real-world Sawyer robot's wrist. In total, we sample 5 trajectories per object per pre-trajectory VSD bin.
%
%
\subsubsection{Results}
%
%
The results for this experiment are illustrated in Figure \ref{fig:experiments:sequential-icp}. In this figure, the graphs on the left and right show the performance of all methods for fusing multiple estimates when using projective Cascading ICP and Hybrid ICP as the underlying ICP algorithms respectively. As these two graphs illustrate, the choice of the underlying ICP algorithm depends on the VSD distribution of the global pose estimator used to initialise sequential ICP. For example, Labbe et al. \cite{CosyPose} report that $63.8\%$ of all VSD errors of the CosyPose pose estimator are below 0.3 for a misalignment tolerance of $\tau=2$ cm. Hence, when used for sequential ICP initialisation, it would be best combined with projective Cascading ICP because in the low pre-trajectory VSD error regime, this performs better than Hybrid ICP. This is because for low pre-ICP VSD errors, projective Cascading ICP has a similar performance to Hybrid ICP while being significantly faster.
\par
Whilst there are some trends in terms of which methods are worst, there is no obvious trend in terms of which method is best. Therefore, we conclude that the Average method is a suitable method for using sequential ICP estimates as it is one of the simplest methods and is at least as good as the others. Also, as methods that used per-image uncertainties did not perform better than other methods, this suggests that either the uncertainty estimates were poor, or these uncertainties were not appropriate for the method proposed by Barfoot et al. \cite{Associating-uncertainty-SE3} for fusing $\textit{SE}(3)$ estimates. Johns \cite{coarse-to-fine-imitation-learning} has also found that methods that used per-image uncertainties were no better than other baselines for fusing sequential pose estimates.

%
%
\section{Conclusions} \label{sec:conclusion}
%
%
We have proposed Hybrid ICP, a flexible ICP variant that uses the current image observation to dynamically choose the optimal data association method and the optimal error metric. As a component of this algorithm, we also introduced Cascading ICP. We showed that Hybrid ICP is more robust to initialisation and noise than other commonly used ICP variants when estimating object poses. We also studied sequential ICP and the trade-off between accuracy and convergence speed and found that for initialisation errors based on state-of-the-art pose estimators, projective Cascading ICP is optimal.
%
%





%
\bibliographystyle{IEEEtran}
\bibliography{IEEEabrv,references}
\end{document}